## Highlights

**Proactive Hierarchical Control Barrier Function-Based Safety Prioritization in Close Human-Robot Interaction Scenarios**

Patanjali Maithani, Aliasghar Arab, Farshad Khorrami, Prashanth Krishnamurthy

- A hierarchical CBF framework with a safety prioritization mechanism that uses a relaxation variable to enable constraint trade-offs based on the criticality of different parts of the human body.

- Validating the method through experimentally testing the proposed technique on Franka Research 3 Robot and use of ZED2i AI camera for human body and pose detection.

# Proactive Hierarchical Control Barrier Function-Based Safety Prioritization in Close Human-Robot Interaction Scenarios

Patanjali Maithani[a], Aliasghar Arab[a,*], Farshad Khorrami[a] and Prashanth Krishnamurthy[a]

[a]*New York University Tandon School Of Engineering, New York City, 11201, New York, USA*

**ABSTRACT**

In collaborative human-robot environments, the unpredictable and dynamic nature of human motion can lead to situations where collisions become unavoidable. In such cases, it is essential for the robotic system to proactively mitigate potential harm through intelligent control strategies. This paper presents a hierarchical control framework based on Control Barrier Functions (CBFs) designed to ensure safe and adaptive operation of autonomous robotic manipulators during close-proximity human-robot interaction. The proposed method introduces a relaxation variable that enables real-time prioritization of safety constraints, allowing the robot to dynamically manage collision risks based on the criticality of different parts of the human body. A secondary constraint mechanism is incorporated to resolve infeasibility by increasing the priority of imminent threats. The framework is experimentally validated on a Franka Research 3 robot equipped with a ZED2i AI camera for real-time human pose and body detection. Experimental results confirm that the CBF-based controller, integrated with depth sensing, facilitates responsive and safe human-robot collaboration, while providing detailed risk analysis and maintaining robust performance in highly dynamic settings.

## 1. Introduction

The integration of autonomous robotic systems into human-centered environments is accelerating in diverse sectors such as manufacturing, healthcare, personal care, and service industries. Among these systems, collaborative robots (cobots) are increasingly deployed to work safely alongside humans, requiring advanced control strategies that ensure both physical and psychological safety without compromising performance. Achieving this balance remains a significant challenge, particularly in close-proximity interaction scenarios where collision risks must be managed in real time. This paper presents a hierarchical control framework based on Control Barrier Functions (CBF) that proactively prioritizes safety constraints using a relaxation variable, enabling the system to adapt to rapidly changing conditions. The framework also incorporates prioritization of more sensitive human body parts and avoidance of infeasible states through an additional constraint mechanism. Experimental validation on a Franka Research 3 robot, combined with high-resolution 3D sensing via a ZED2i AI camera for real-time human pose detection, demonstrates the controller's ability to proactively mitigate collision risks and support seamless, safe collaboration in dynamic environments.

We introduce a novel control approach that accounts for the varying vulnerability of different parts of the human body. We implement a hierarchical CBF framework to ensure safe control of the robot, explicitly addressing the higher risk of injury to sensitive areas during robot-human interactions. The compliant controller incorporates a potential injury analysis, establishing the safe distance and velocity between human and robot to avoid any impact that might cause injury. This design prioritizes enhanced safety for more vulnerable regions, minimizing the risk of injury with higher safety margins. For example, the head is a higher priority than the hand, and safety factors are considered more seriously, as shown in Fig. 1. The hierarchical approach proposed in this paper can be integrated into any existing robot controller to impose safety compliance, unlike our recent article, which integrated nonlinear model predictive control (NMPC) with feedback linearization for smooth direct control of robot commands Arab, Mousavi, Yu, and Kucukdemiral (2024).

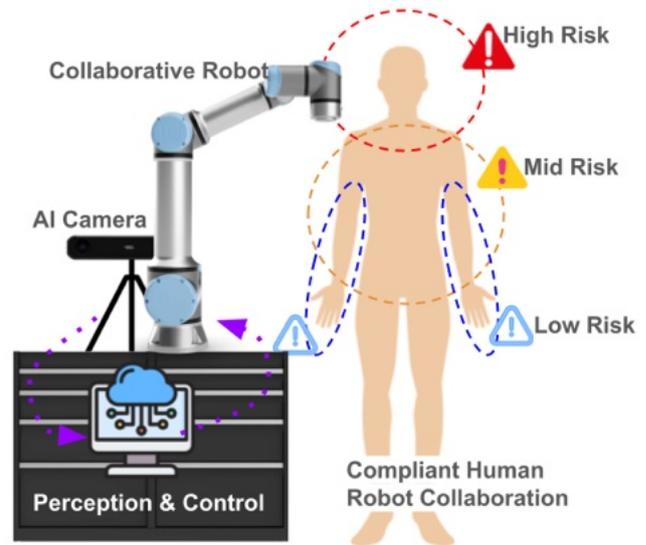

**Figure 1:** A schematic of a human robot setup in a collaborative environment and risk levels for different part of the body.

P. Maithani, and A. Arab are with the Department of Mechanical and Aerospace Engineering at New York University, Brooklyn, NY, USA (emails: aliasghar.arab@nyu.edu, pm3516@nyu.edu), F. Khorrami and P. Krishnamurthy are with the Department of Electrical Engineering at New York University, Brooklyn, NY, USA (emails: khorrami@nyu.edu, prashanth.krishnamurthy@nyu.edu).

ORCID(s):

The design of safety-guaranteed control strategies to ensure safe human-robot collaboration can accelerate the acceptance of sociotechnological systems such as cobots in various domains to improve work efficiency and performance and fill workforce shortages. Unlike traditional industrial robotic arms, which operate within protective cages, collaborative robots are designed for direct physical interaction, sharing a common workspace with humans, and performing tasks together as a teammate Villani, Pini, Leali, and Secchi (2018); Sun, Peng, Zou, Zhou, and Wu (2025). This requires reliable control systems capable of mitigating the risk of collision in any scenario in unstructured environments. Recent literature highlights CBFs as a promising solution to this challenge Ferraguti, Talignani Landi, Costi, Bonfé, Farsoni, Secchi, and Fantuzzi (2019); Jian, Yan, Lei, Lu, Lan, Wang, and Liang (2023); Singletary, Guffey, Molnar, Sinnet, and Ames (2022). CBFs can be conceptualized as mathematical tools that constrain the control actions of a dynamical system, enabling forward invariance for a desired set of states Ames, Coogan, Egerstedt, Notomista, Sreenath, and Tabuada (2019). This desired state set can either keep the robot away from obstacles or compel the system to exhibit specific behaviors, such as human follow-up or occlusion-free movement for robotic manipulators. Such manipulators must continuously adjust their maneuvers to maintain a clear view of their targets, regardless of any obstructions in their field of sight Pupa, Breveglieri, and Secchi (2023); Wei, Dai, Khorrambakht, Krishnamurthy, and Khorrami (2024).

A general approach to forming safety constraints using CBFs involves differentiating the CBF to establish a linear relationship with the control input variable. In some cases, multiple differentiations are necessary to achieve this linearity; these CBFs are referred to as High-Order Control Barrier Functions (HOCBFs) Shi and Hu (2024). In the context of robotic manipulators, HOCBFs facilitate torque-based control Shi and Hu (2023); Zhang, Li, and Figueroa (2024), while velocity-based controllers typically utilize a single time derivative of the CBF to derive a linear relationship with the control input. Despite their potential, the practical implementation of CBFs for obstacle avoidance and ensuring human safety faces several challenges. These include the requirement that CBFs must be smooth, the need for accurate state estimates of both the obstacles and the robotic system itself, and the lack of a prioritization framework for higher-risk events.

Liu, Li, Duan, Wu, Wang, Fan, Lin, and Hu (2023) proposed a hierarchical inverse dynamics control strategy for robot manipulators that neglect human-robot interactions. In human-robot interactions, the distances between robot links and humans can be modeled as CBFs, leading to solutions for obstacle avoidance based on this principle. Signed Distance Functions (SDFs) have also been examined as CBFs in various studies. To qualify as a CBF, a function must be continuous throughout the state space; however, SDFs may not meet this criterion in certain edge cases, but still demonstrate satisfactory performance Singletary et al. (2022). Dai, Khorrambakht, Krishnamurthy, Goncalves, Tzes, and Khorrami (2023a) have employed a differentiable collision detection technique to develop a continuous CBF, although this approach increases computational time. Furthermore, the dynamic nature of interactions between robotic manipulators and human workers requires the use of time-varying CBFs, where the time derivative of the obstacle pose in the defined CBF must also be considered when formulating the safety constraints within a quadratic programming framework (QP) Dai, Khorrambakht, Krishnamurthy, and Khorrami (2023b); Tracy, Howell, and Manchester (2023). In addition to these considerations, other constraints—including joint limits, joint velocity limits, and singularity avoidance—are crucial to designing safe and performance-centric control policies for robotic manipulators Zhang et al. (2024). Hence, a multiobjective constrained optimization problem should be solved in real time to impose limits and comply with standards and safety regulations Ferraguti, Bertuletti, Landi, Bonfè, Fantuzzi, and Secchi (2020). Liu, Lyu, Guo, Wang, Yu, and Guo (2024) developed a coordinated control framework for aerial manipulators to enable safe and compliant physical interactions during contact tasks, whose approach may face challenges in dynamic or unpredictable environments such as human interactions.

A proactive strategy to improve robustness in dynamic environments is to predict the trajectories of future obstacles and incorporate these data into safety constraints Kim, Lee, and Ames (2024). A similar approach is used by Dai et al. (2023b), where robot links and obstacles are represented as convex shapes, with collision detection performed by checking for intersections. In the corresponding QP optimization problem, multiple constraints are considered, such as obstacle avoidance, joint limits, and joint velocity limits, but conflicts among constraints can lead to infeasibility Isaly, Edwards, Bell, and Dixon (2023) compromising human safety. Although these edge-case scenarios are rare, the QP solver may fail in providing a solution Palmieri, Lillo, Lippi, Chiaverini, and Marino (2025). In Gonçalves, Krishnamurthy, Tzes, and Khorrami (2024), a circulation constraint was added to reduce the number of spurious equilibra. The literature suggests that in such cases, the constraints associated with human safety should be prioritized over others Arab (2023); Notomista, Mayya, Selvaggio, Santos, and Secchi (2020). To address this, relaxation variables can be introduced for constraints of lower priority Lee, Kim, and Ames (2023). However, these relaxation techniques are still being developed and more experimental validation is required to assess their effectiveness.

The main objective of this paper is to design and validate a controller for a collaborative robot arm operating in close proximity to a human, as illustrated in Fig. 1. Thus taking the concerns into account discussed above, this paper implements a CBF based approach for safety enhanced control of a collaborative robot manipulator operating alongside human worker through a novel prioritization mechanism. Experimental results demonstrate how the cobot safely and smoothly avoids human's body(hand and head). Safety prioritization is capable of ensuring that the highest priority constraints will not be neglected at any price. The main contributions of the proposed safe control approach for collaborative robots are listed below.

1. A hierarchical CBF framework with a safety prioritization mechanism that uses a relaxation variable to enable constraint trade-offs based on the criticality of different parts of the human body.
2. Exeprimental validation of the proposed technique on Franka Research 3 Robot and use of ZED2i AI camera for human body and pose detection.

The remainder of the paper is structured as follows. Section II explains the problem and details the associated technical challenges. Section III explains the methodology for safe control based on barrier functions of control. Section IV presents results that validate the safety set invariance for the manipulator operating in close proximity of a moving human collaborator. Finally, Section V concludes the paper and discusses avenues for future work.

*Notation*: The following notations are used throughout the paper. Lowercase bold letters denote vectors, e.g., $\boldsymbol{x} = [x_1, x_2, \ldots, x_n]^\top \in \mathbb{R}^n$, while uppercase bold letters denote matrices, e.g., $\boldsymbol{A} \in \mathbb{R}^{m \times n}$. The origin of the vector space with the appropriate dimension is denoted by $\boldsymbol{0}$, and $\mathrm{diag}(\boldsymbol{v})$ represents a diagonal matrix with diagonal elements of the vector $\boldsymbol{v}$. The L2 norm of a vector $\boldsymbol{x}$ is denoted as $\|\boldsymbol{x}\|_2 \triangleq (\boldsymbol{x}^\top \boldsymbol{x})^{1/2}$. A continuously differentiable function, often used as a CBF, is denoted by $h(\boldsymbol{x})$, and its gradient with respect to $\boldsymbol{x}$ is given by $\frac{\partial h(\boldsymbol{x})}{\partial \boldsymbol{x}} = \left[\frac{\partial h(\boldsymbol{x})}{\partial x_1}, \frac{\partial h(\boldsymbol{x})}{\partial x_2}, \ldots, \frac{\partial h(\boldsymbol{x})}{\partial x_n}\right]$. The time derivative of $h(\boldsymbol{x})$ is denoted by $\frac{dh(\boldsymbol{x})}{dt}$. The state vector of the system is represented by $\boldsymbol{x} \in \mathbb{R}^n$, and the control input vector is denoted by $\boldsymbol{u} \in \mathbb{R}^m$. The joint states and joint velocities of the robotic manipulator are represented by $\boldsymbol{q} \in \mathbb{R}^n$ and $\dot{\boldsymbol{q}} \in \mathbb{R}^n$, respectively.

The desired and actual positions of the end-effector are denoted by $\boldsymbol{p}_d \in \mathbb{R}^3$ and $\boldsymbol{p}_{\mathrm{actual}} \in \mathbb{R}^3$, respectively, while the position of an obstacle (e.g., human head or hand) is represented by $\boldsymbol{p}_{\mathrm{obs}} \in \mathbb{R}^3$. The positional part of the Jacobian matrix of the end effector is denoted by $J_{\mathrm{ee}} \in \mathbb{R}^{3 \times n}$, and its pseudo-inverse is represented by $J_{\mathrm{ee}}^\dagger$. The projection matrix for the null space is denoted by $J_{\mathrm{null}} \in \mathbb{R}^{n \times n}$. The safe set of states is denoted by $\mathcal{X}_s \subset \mathbb{R}^n$, while the unsafe set of states is represented by $\mathcal{X}_u \subset \mathbb{R}^n$. The reachable set of states is denoted by $\mathcal{D} \subset \mathbb{R}^n$, and the set of admissible control inputs is represented by $\mathcal{U} \subset \mathbb{R}^m$. The joint velocity commands generated by the performance controller and adjusted by the safety controller are indicated by $\dot{\boldsymbol{q}}_{\mathrm{perf}} \in \mathbb{R}^n$ and $\dot{\boldsymbol{q}}_{\mathrm{safe}} \in \mathbb{R}^n$, respectively. The relaxation variable for the hand constraint in the optimization problem is indicated by $\delta_{\mathrm{ha}} \in \mathbb{R}$, while the penalty parameter and the parameter influencing the behavior of the control barrier function are represented by $\beta \in \mathbb{R}$ and $\gamma \in \mathbb{R}$, respectively. The time variable is denoted by $t$, and the gain parameter in the performance controller is represented by $\lambda > 0$. An extended class $\boldsymbol{K}$ function used in the control barrier function framework is indicated by $\alpha(h(\boldsymbol{x}))$, and a safety parameter to ensure collision avoidance is represented by $\epsilon$. The radii of spheres generalized around the robot link and the human body part are denoted by $R_j$ and $R_i$, respectively. In the results section, the relative velocity between the end-effector and the human head or hand is denoted by $v_{\mathrm{rel}}$, calculated as $v_{\mathrm{rel}} = \frac{(\boldsymbol{p}_{\mathrm{ee}} - \boldsymbol{p}_h)^\top (\boldsymbol{v}_{\mathrm{ee}} - \boldsymbol{v}_h)}{\|\boldsymbol{p}_{\mathrm{ee}} - \boldsymbol{p}_h\|}$, where $h$ represents either the head or the hand. The distances from the end-effector to the head and hand are denoted by $d_{\mathrm{head}}$ and $d_{\mathrm{hand}}$, respectively, and the maximum relaxation variable value reached during experiments is denoted by $\delta_{\mathrm{max}}$.

## 2. Problem Formulation

A collaborative robot operating in close proximity to a human—illustrated in Fig. 2—can be formulated as a non-linear control problem. In such settings, unpredictable and dynamic human motion introduces scenarios where collisions may become unavoidable, requiring the robot to proactively mitigate risk through adaptive control. To achieve this, the robot uses a depth camera to perceive and track the relative position and movement of the human in its local frame. The robot dynamics, along with safety constraints derived from human body-part proximity, are modeled to enable prioritized and flexible collision avoidance using a hierarchical control framework based on CBFs as

$$\Sigma : \quad \dot{\boldsymbol{x}} = f(\boldsymbol{x}) + g(\boldsymbol{x})\boldsymbol{u} \qquad (1)$$
$$\text{s.t. } \boldsymbol{u} \in \mathcal{U}$$
$$\boldsymbol{x} \in \mathcal{X}_s$$

where $f : \mathbb{R}^n \to \mathbb{R}^n$ and $g : \mathbb{R}^n \to \mathbb{R}^{n \times m}$ represents the dynamical system of the robot as a non-linear function and $\boldsymbol{x} \in \mathcal{D} \subset R^n$ is the state and $\boldsymbol{u} \in \mathcal{U} \subseteq R^m$ the input of the system, where $n$ and $m$ are the state and the length of the input vector. As we aim to make a velocity-based controller for the robotic manipulator, the dynamics for the system considered is $\dot{\boldsymbol{x}} = g(\boldsymbol{x})\boldsymbol{u}$, $f(\boldsymbol{x}) = O_{7 \times 1}$, $g(\boldsymbol{x}) = I_{7 \times 7}$, where $\boldsymbol{x}$ represents the joint states $\boldsymbol{q}$, of the manipulator and $\boldsymbol{u}$ corresponds to the joint velocity control signal. The key idea behind using CBFs for dynamic obstacle avoidance is to define a time-varying safe set in the state space that the robot

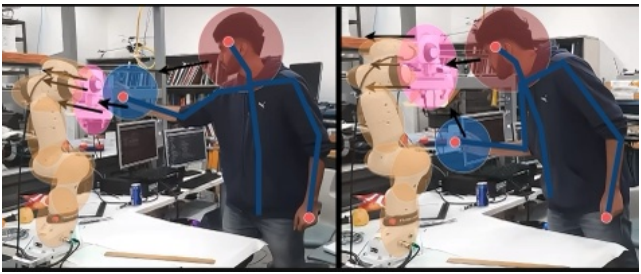

**Figure 2:** Images showing extracted body framework and barriers for the proactive robot controller.

CEP draft-1

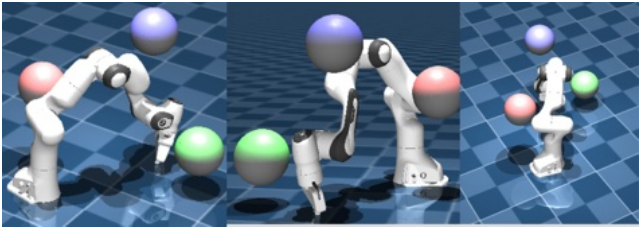

**Figure 3:** Scenario where solver has no solution due to conflict among the constraints found with numerical coverage test.

must stay within to avoid collisions. The controller intends to operate the robot states invariant to the safe set $\mathcal{X}_s \subset \mathbb{R}^n$ by choosing the controller commands from the admissible sets $\mathcal{U}$. This safe set can be defined implicitly as the 0-superlevel set of a continuously differentiable function to form barrier functions. These functions are parametrized to conform to the human motion around the robot using relative distances and motions.

**Assumption 1** (Safe Forward Invariance). *The control objective is to determine an input $u \in \mathcal{U}$ such that the nonlinear system state $x(t)$ converges to a desired target $x_d(t)$ while remaining within a forward-invariant safe set $\mathcal{X}_s \subset \mathcal{D}$ for all $t \geq 0$. The safe set is defined as*

$$\mathcal{X}_s = \mathcal{D} \setminus \left\{ \bigcup_{i \in \mathcal{I}} \mathcal{X}_u^i \right\} \qquad (2)$$

*where each $\mathcal{X}_u^i$ denotes an unsafe subset associated with high-level safety rules indexed by $i \in \mathcal{I}$.*

**Lemma 1** (Existence of Infeasible Safe Commands Without Prioritization). *There exists a configuration of unsafe sets $\mathcal{X}_u^i$ such that, for a given system state $x(t) \in \mathcal{X}_s$, no admissible control input $u \in \mathcal{U}$ can be found that ensures $x(t + \Delta t) \in \mathcal{X}_s$, i.e., the system must inevitably violate at least one safety constraint.*

*Proof.* Consider a scenario in which the robot state $x(t) \in \mathcal{X}_s$ is surrounded by multiple unsafe regions $\mathcal{X}_u^i$, $i \in \mathcal{I}$, such that every admissible control input $u \in \mathcal{U}$ results in a trajectory intersecting at least one of the unsafe sets, i.e.,

$$\forall u \in \mathcal{U}, \quad \exists t' > t \text{ such that } x(t') \in \mathcal{X}_u^i \text{ for some } i \in \mathcal{I}. \qquad (3)$$

This implies the set of admissible control inputs that preserve forward invariance of $\mathcal{X}_s$ is empty:

$$\{ u \in \mathcal{U} \mid x(t + \Delta t) \in \mathcal{X}_s \} = \emptyset. \qquad (4)$$

In such cases, the optimization problem becomes infeasible, and the system must enter at least one unsafe region, such are the scenarios shown in the simulations in Fig. 3. Without a prioritization mechanism, the controller cannot distinguish between high- and low-risk constraint violations. Thus, the need for a prioritization scheme (e.g., hierarchical relaxation) arises to manage safety trade-offs and maintain feasibility when constraints are in conflict. □

A human is a dynamic object, so for a collaborative robot, the unsafe zones are also dynamic and may shift during the robot's operation. The safe region in green indicates the states that we are allowed to operate the robot, while the unsafe region highlighted with orange and red is unsafe set, where red demonstrates higher risk areas. An orange area is moving and puts the robot in a situation that a collision is inevitable, hence the robot must choose to hit the lower risk part of the body, as shown in Fig. 4.

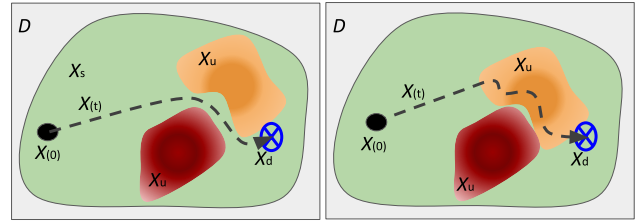

**Figure 4:** A schematic of robot operation in close proximity of human body where head ($1^{st}$ priority) is the red area and hand ($2^{nd}$ priority) is the orange area and the black line shows the robot trajectory.

The prioritization mechanism will put higher-risk unsafe regions at top priority constraints to avoid, such as prioritizing head over hand in a human avoidance scenario in a collaborative robot working in close proximity to a human. By establishing safety constraints within a hierarchical control architecture that can be integrated with existing robot controllers, we aim to ensure that the robot achieves its target while prioritizing obstacle avoidance based on the injury risk associated with different parts of the human body. To accomplish this, we utilize a 3D depth camera to detect and track the human body, allowing the robot to adapt its movements dynamically. The control approach is tested on a collaborative robot arm to validate safety and effectiveness. The key challenges in developing an effective risk-aware collision avoidance framework for human-robot collaboration applications are

- Accurate and reliable sensing of the human within the robot environment, including precise detection of the positions of various parts of the body.
- A safety-assured control policy that prevents potential collisions, with robot motion constraints continuously prioritized based on the risk of harm, maintaining adherence to safety limits throughout the operation.

## 3. Hierarchical Safe Control Design

In our hierarchical safe control design, Fig. 5, we separate performance and safety responsibilities into distinct layers. In the first layer, a performance controller calculates the velocity of each robot joint to follow the desired trajectory. In the second layer, the safety controller independently monitors the state of the robot, ensuring that any velocity commands remain within a safe subset of possible states. This hierarchical structure allows the performance controller to operate unimpeded while the safety controller continuously enforces stability and constraint adherence. Consequently, even if the performance controller generates commands that could theoretically violate safety boundaries, the inner-layer safety mechanism overrides or adjusts those commands in real time to maintain safety guarantees with an additional prioritization mechanism explained in this section.

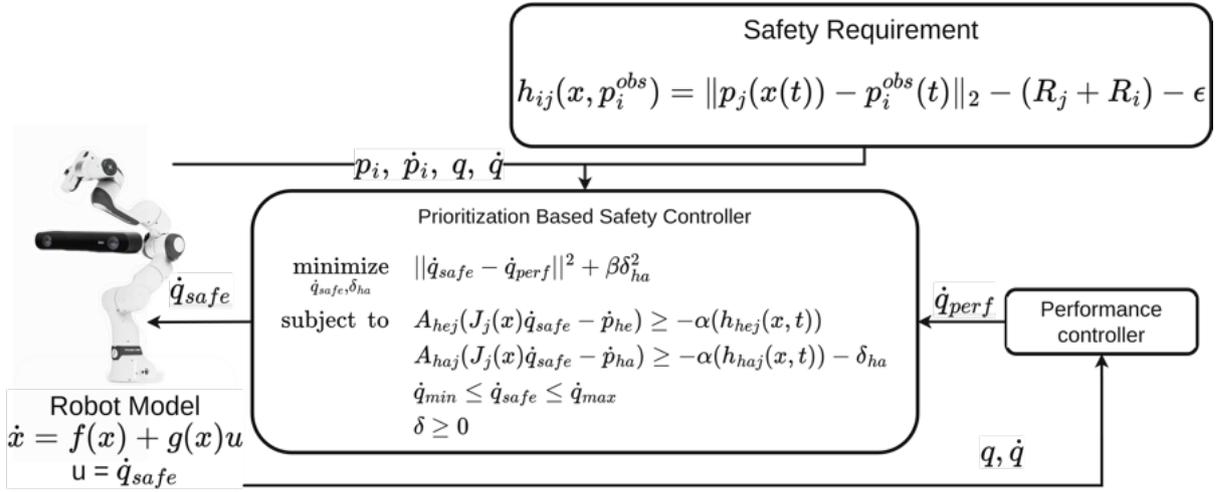

**Figure 5:** Overall architecture of the proposed CBF-based safety.

### 3.1. Performance Controller

As mentioned earlier, the main objective of the performance controller is to compute the desired joint velocity commands that make the end-effector trace the desired 3D trajectory. The following inverse kinematics controller is used to achieve this:

$$\dot{q}_{\text{perf}} = J_{\text{ee}}^{\dagger}\left(\dot{p}_d - \lambda(p_{\text{actual}} - p_d)\right) + \dot{q}_{\text{mid}}, \quad \lambda > 0 \quad (5)$$

where, $p_d$, $\dot{p}_d$, and $p_{\text{actual}}$ correspond to the desired position, the desired velocity, and the actual position of the end effector, respectively. $J_{\text{ee}}^{\dagger}$ represents the pseudo-inverse of the positional part of the end-effector Jacobian as

$$J_{\text{ee}}^{\dagger} = J_{\text{ee}}^T \cdot \left(J_{\text{ee}} \cdot J_{\text{ee}}^T + 0.01 \cdot I\right)^{-1} \quad (6)$$

with $\dot{q}_{\text{mid}}$ being defined as

$$\dot{q}_{\text{mid}} = -J_{\text{null}} \cdot \left(K_p^{\text{joint}} \cdot e_{\text{joint}}\right), \quad (7)$$

$K_p^{\text{joint}}$ is the gain matrix and $J_{\text{null}}$ represents the null-space projection matrix given as

$$J_{\text{null}} = I - J_{\text{ee}}^{\dagger} \cdot J_{\text{ee}}. \quad (8)$$

The joint error is $e_{\text{joint}} = W \cdot (q - \bar{q})$. The quantities $W$ and $\bar{q}$ are defined as $W = \text{diag}\left(\frac{1}{j_{\text{ub}}-j_{\text{lb}}}\right)$ and $\bar{q} = \frac{1}{2} \cdot (j_{\text{ub}} + j_{\text{lb}})$ with $j_{\text{ub}}$ and $j_{\text{lb}}$ representing the upper and lower bounds of the joints. The component $\dot{q}_{\text{mid}}$ helps to bring the joint configurations closer to their mid-joint values without affecting the desired linear velocity of the end effector.

### 3.2. Control Barrier Function Design

For the dynamical system defined in (1), the safe set $\mathcal{X}_s$ is implicitly defined as the 0-superlevel set of a continuously differentiable function $h(x, p^{obs})$:

$$S = x : h(x, p^{obs}) \geq 0 \quad (9)$$

where the colliding states are given by the set $x : h(x, p^{obs}) < 0$. The function $h(x, p^{obs})$ is called the control barrier function if $\frac{\partial h}{\partial x}(x, p^{obs}) \neq 0$ for all $x$ within the bound of $S$ and there exists an extended class $K$ function $\alpha$ such that for the control system (1) and for all $x \in S$

$$\frac{dh(x, p^{obs})}{dt} \geq -\alpha(h(x, p^{obs})). \quad (10)$$

If the above inequality is enforced at all times, then it can be guaranteed that $x$ will remain in the set $S$. An extended class $K$ function, $\alpha$, is a function that increases strictly in its domain, and $\alpha(\mathbf{0}) = 0$. In our case, the distance between the robot's different links and the operator's head and hand must always be greater than a certain threshold. Therefore, the CBF can be expressed as

$$h_{ij}(x, p_i^{obs}) = \|p_j(x) - p_i^{obs}(t)\|_2 - (R_j + R_i) - \epsilon, \quad (11)$$

$$j = 1, 2, \ldots, n$$

where $p_j$ is the position vector of the midpoint of the robot link $j$ (robot with $n$ links) and $p_i$ is the position of the operator's head / hand ($i = head/hand$). $R_j$ and $R_i$ are the radii for the spheres generalized around the link's midpoint and head/hand of the operator. Epsilon, $\epsilon$, is the safety parameter to ensure collision avoidance. $\|.\|_2$ is the $L2$ norm which corresponds to the square root of the sum of squares of the differences between the position vectors between the robot's link and head/hand of the operator. To form the inequality present in (10), $h(x, p_i^{obs})$ term needs to be differentiated with respect to time:

$$\frac{dh(x, p_i^{obs})}{dt} = \frac{\partial h(x, p_i^{obs})}{\partial x}\dot{x} + \frac{\partial h(x, p_i^{obs})}{\partial t}$$
$$= \frac{\partial h(x, p_i^{obs})}{\partial x}(f(x) + g(x)u) + \frac{\partial h(x, p_i^{obs})}{\partial p_i^{obs}}\frac{\partial p_i^{obs}}{\partial t} \quad (12)$$

. As introduced in the Problem Formulation, we consider a velocity-based control model with simplified dynamics: $\dot{x} = g(x)u$, $f(x) = O_{7\times 1}$, $g(x) = I_{7\times 7}$, $x = q$, where $q$ denotes the joint states and $u$ the joint velocity inputs. After doing the further derivation from Eq. (12), the expression for the L.H.S in Eq. (10) turns out to be

$$\frac{dh_{ij}(x, p_i^{obs})}{dt} = A_{ij}(J_j(x)u - \dot{p}_i)$$
$$\text{where } A_{ij} = \frac{(\mathbf{p}_j - \mathbf{p}_i)^T}{\sqrt{(\mathbf{p}_j - \mathbf{p}_i)^T(\mathbf{p}_j - \mathbf{p}_i)}} \quad (13)$$
$$j = 1, 2, \ldots, n$$

where $J_j(x)$ is the 3×7 Jacobian matrix for the j-th robot link. Suppose a performance control policy $u = \dot{q}_{perf}$ is given to the manipulator, then to ensure that it is in compliance with the inequality in (10), an optimization control problem (OCP) is formed as

$$\min_{\dot{q}_{safe}} ||\dot{q}_{safe} - \dot{q}_{perf}||^2 \quad (14)$$
$$\text{s.t. } (i = \text{head, hand})$$
$$A_{ij}(J_j(x)\dot{q}_{safe} - \dot{p}_i) \geq -\alpha(h_{ij}(x, p_i^{obs})),$$
$$\dot{q}_{min} \leq \dot{q}_{safe} \leq \dot{q}_{max},$$
$$j = 1, 2, \ldots, n$$

where, the formulated quadratic problem in Eq. 14 may experience conflicts between constraints, leading to infeasibility. Therefore, we adopt a relaxed form of the quadratic program (QP) by introducing a relaxation variable corresponding to the hand constraint. This adjustment modifies Eq. 14 as follows

$$\min_{(\dot{q}_{safe}, \delta_{ha} \geq 0)} \left\{ ||\dot{q}_{safe} - \dot{q}_{perf}||^2 + \beta \delta_{ha}^2 \right\} \quad (15)$$
$$\text{s.t.}$$
$$A_{hej}(J_j(x)\dot{q}_{safe} - \dot{p}_{he}) \geq -\alpha(h_{hej}(x, p_{he}^{obs})),$$
$$A_{haj}(J_j(x)\dot{q}_{safe} - \dot{p}_{ha}) \geq -\alpha(h_{haj}(x, p_{he}^{obs})) - \delta_{ha},$$
$$\dot{q}_{min} \leq \dot{q}_{safe} \leq \dot{q}_{max},$$
$$j = 1, 2, \ldots, n$$

where $he$ = head, $ha$ = hand, and $\alpha(h_{ij}(x, p_i^{obs})) = \gamma h_{ij}(x, p_i^{obs})$ ($\gamma > 0$). The parameter $\gamma$ significantly influences the system's behavior. When $\gamma$ is set to a lower value, the control signal ensures that the system states remain conservatively within the safe set, leading to a larger deviation from the desired path. In contrast, when $\gamma$ is increased, the control signal enables the states of the system to operate closer to the boundary of the safe set, allowing for a smaller deviation from the desired path. This adjustment of $\gamma$ reflects a trade-off between safety and path adherence, where lower values prioritize safety at the cost of straying from the intended trajectory, while higher values allow for more direct tracking of the desired path but can increase the risk of approaching unsafe conditions. By incorporating a relaxation variable into the hand constraint, we assign it a lower priority compared to the head avoidance constraint.

We now state two propositions about feasibility and convergence.

**Proposition 1** (Feasibility Preservation). *If the original problem (14) is feasible, then the relaxed problem (15) is also feasible for any $\beta > 0$.*

*Proof.* Let $\dot{q}^\star$ be a feasible point for (14), i.e.,

$$A_{hej}(J_j(x)\dot{q}^\star - \dot{p}_{he}) \geq -\alpha(h_{hej}(x, p_{he}^{obs})),$$
$$A_{haj}(J_j(x)\dot{q}^\star - \dot{p}_{ha}) \geq -\alpha(h_{haj}(x, p_{ha}^{obs})).$$

where $\delta_{ha} = 0$. Then, $(\dot{q}^\star, 0)$ satisfies all constraints in (15). Hence, the relaxed problem is feasible whenever the original problem is feasible. □

**Proposition 2** (Convergence to Original Solution). *Assume the original QP (14) is feasible and let $(\dot{q}^\beta, \delta_{ha}^\beta)$ denote the optimal solution to (15) for a given $\beta > 0$. Then:*

$$\lim_{\beta \to \infty} \delta_{ha}^\beta = 0, \quad \text{and} \quad \lim_{\beta \to \infty} \dot{q}^\beta = \dot{q}^\star,$$

*where $\dot{q}^\star$ is the unique solution to (14). In numerical implementations, $\delta_{ha}^b$ may remain within a small neighborhood of zero due to solver tolerances.*

*Proof.* The cost function of (15) is strictly convex in $(\dot{q}, \delta_{ha})$, so the solution is unique for all $\beta > 0$. Define the optimal value function:

$$V(\beta) := \min_{(\dot{q}, \delta_{ha} \geq 0)} \left\{ ||\dot{q} - \dot{q}_{\text{perf}}||^2 + \beta \delta_{ha}^2 \text{ s.t. constraints} \right\}.$$

For any feasible $\dot{q}$ such that the unrelaxed constraints are satisfied, we can choose $\delta_{ha} = 0$(for $(\dot{q}^\star, 0)$), which yields:

$$||\dot{q}^\star - \dot{q}_{\text{perf}}||^2 + \beta \cdot 0^2 = ||\dot{q}^\star - \dot{q}_{\text{perf}}||^2.$$

Since $V(\beta)$ is defined as the minimum of the objective function over the feasible set, and we have evaluated the objective at a feasible point, it must hold that:

$$V(\beta) \leq ||\dot{q}^\star - \dot{q}_{\text{perf}}||^2.$$

where $\lim_{\beta \to \infty} \delta_{ha}^\beta = \varepsilon > 0$. Then the cost satisfies:

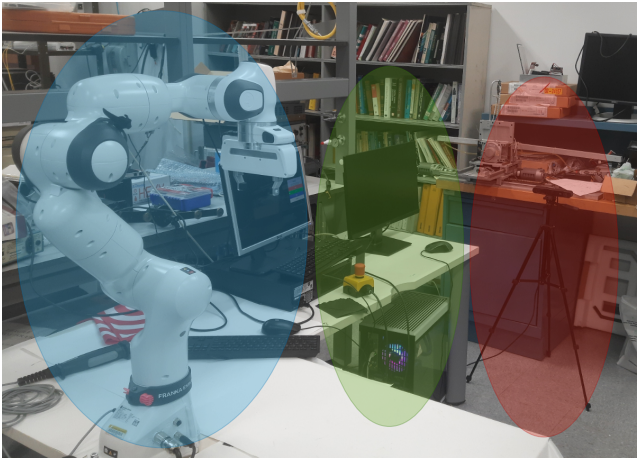

**Figure 6:** The blue region indicates the Franka robot, the red region highlights the ZED 2i camera, and the green region shows the computing system.

offers high-resolution depth perception and stereo vision capabilities. Our setup incorporated a 34-keypoint model operating at an inference rate of 60 Hz, allowing for real-time human pose estimation and responsive interaction with the manipulator. Before the experiments, we performed an eye-on-base calibration to calculate the transformation matrix between the manipulator's base and the depth camera. This comprehensive framework facilitated a thorough evaluation of the system's performance and safety in various human-robot interaction scenarios.

For the validation of our approach, barrier functions between the end-effector and the human head and hand have been considered. The end-effector Jacobian of the manipulator was computed using the Pinocchio library, while the ProxSuite library was utilized to solve the optimization problem outlined in Eq. (15), where $j$=end-effector of the robot Bambade, Schramm, Kazdadi, Caron, Taylor, and Carpentier (2023). The performance controller given in Eq. (5) is used.

$$V(\beta) = \|\dot{q}^\beta - \dot{q}_{\text{perf}}\|^2 + \beta(\delta_{ha}^\beta)^2 \geq C + \beta\varepsilon^2 \to \infty,$$

which contradicts the bounded upper estimate. Hence, $\delta_{ha}^\beta \to 0$.

Let $\mathcal{F}_\beta$ be the feasible set of the relaxed QP. Since the feasible sets $\mathcal{F}_\beta$ converge to the feasible set of the original problem as $\delta_{ha}^\beta \to 0$, and the cost is continuous and convex, it follows by uniqueness and continuity of minimizers that $\dot{q}^\beta \to \dot{q}^\star$. In practice, due to numerical tolerances and finite-precision arithmetic, the computed relaxation variable may not be identically zero but will lie within a small neighborhood of zero. □

Here, the parameter $\beta$ plays a crucial role in determining the degree to which the relaxation variable $\delta_{ha}$ achieves a non-zero value. When $\beta$ is large, the quadratic penalty term $\beta\delta_{ha}^2$ dominates the objective, discouraging unnecessary relaxation of the hand constraint. As a result, $\delta_{ha}$ remains small unless strictly required to resolve conflicts between constraints, thus ensuring the constraint is satisfied to the greatest extent possible. In contrast, a small $\beta$ reduces the influence of this penalty, allowing a greater relaxation of the hand constraint to better track the desired velocity $\dot{q}_{\text{perf}}$. In such cases, optimization prioritizes performance over strict safety, which may lead to violations involving the lower-priority body part (hand). Thus, when $\beta$ is large, the relaxation variable allows a *best effort* satisfaction of the hand constraint while preserving strict adherence to the head constraint, highlighting the importance of carefully tuning $\beta$ to balance safety and performance.

## 4. Experimental Results

The proposed control policy was implemented on a 7-DOF manipulator, Franka Research 3, Fig. 6. For human pose detection, we used the ZED2i 3D depth camera, which

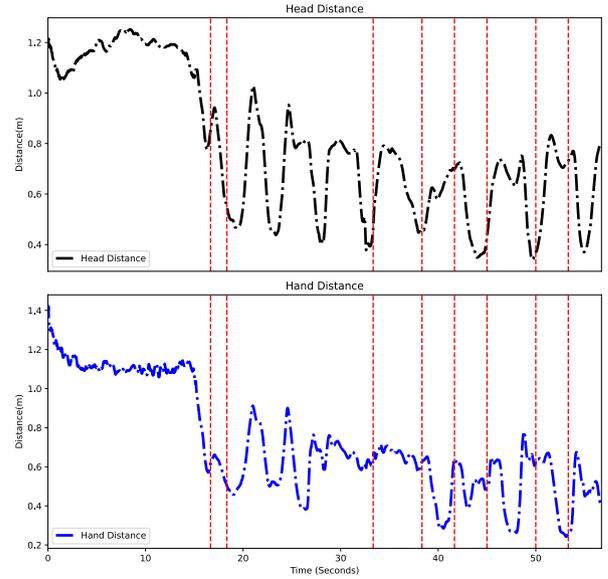

**Figure 7:** Head and hand distances from the end-effector.

Initially, the controller was tested in a scenario in which the operator was not in proximity to the manipulator. The red trajectories depicted in Figs. 9 and 8 correspond to this case. Subsequently, an operator was placed in proximity with the manipulator and instructed to vary the position of their head and hand randomly. The resulting response is illustrated by the black trajectory in the same figures.

From Figs. 9 and 8, we conclude that the major deviations in the joint trajectories are primarily attributed to joints 1, 3, and 5 when the desired end-effector position is at rest(from t=42 seconds). Including Fig. 7, which shows the distances from the operator's head and hand to the end-effector, it is clearly demonstrated that the manipulator quickly deviates from its nominal trajectory when the operator's head or hand enters its vicinity. Subsequently, when the operator is away, the manipulator retracts back to the nominal trajectory, as defined by the performance controller. This behavior illustrates the system's efficiency and safety in human-robot interaction scenarios. For further validation of our proposed approach, the controller was tested in different scenarios as shown in Table 1. Table 1 shows the average values for the relative velocities and relative distances of the head and hand with respect to the end effector for the instances where the distance is less than 0.6 meters or the relative velocity is less than 0 $m/s$ (scenario when the obstacle and end effector are moving towards each other). The expression to compute the relative velocity is

$$v_{rel} = \frac{(p_{ee} - p_h)^T (v_{ee} - v_h)}{||p_{ee} - p_h||}, h = head, hand. \quad (16)$$

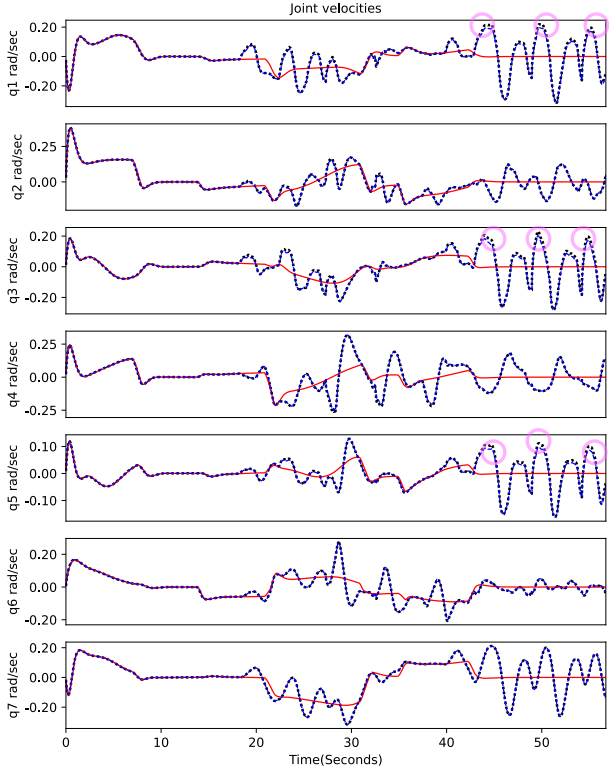

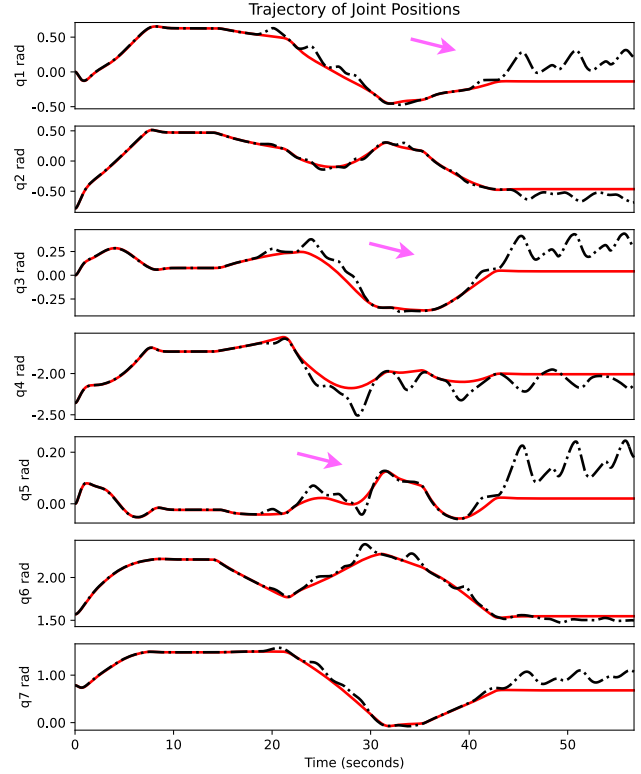

**Figure 8:** Joint velocities with relaxed hand constraint. Blue: performance (Eq. 5); Red: no obstacle; Black: safe control. Pink circles show key deviations from performance.

**Figure 9:** Joint positions with relaxed hand constraint. Red: no-obstacle case; Black dash-dotted: head and hand near end-effector. Pink arrows show major trajectory deviations due to Joints 1, 3, and 5 after $t = 42$ s.

Based on experimental data, Fig. 10 illustrates the prioritization behavior: at close distances, the area with negative relative velocity is larger for the hand than for the head, indicating a more aggressive hand motion. When the hand approaches the end-effector rapidly, the system reacts quickly; slower approaches may result in contact. In contrast, head approaches consistently trigger avoidance, demonstrating the effect of the relaxation variable. This comes from the prioritization weight $\beta$ associated with $\delta_{max}$, which favors task tracking while allowing lower-priority constraints (e.g., hand) to be relaxed—ideal for collaborative interaction. Although the current set-up involved only the hand and head, future work will explore adaptive tuning of $\beta$ based on environmental context. For example, rapid hand movement can signal danger (high $\beta$), while slow and deliberate motion could reflect collaboration (low $\beta$). In tasks involving sharp tools, even slow approaches should increase $\beta$ to ensure safety. Although rule-based strategies may suffice in simple cases, learning-based methods to adaptively adjust $\beta$ in complex environments with multiple obstacles and robot links present a promising research direction.

### 4.1. Highlighting Prioritization and Coverage Test

To further highlight the effect of the relaxation variable on the proposed control approach, a simulation is also performed on a 2-DOF manipulator. The scenario considers two spherical obstacles (red and green) approaching the end-effector of the manipulator, as shown in Fig. 11. The red obstacle is assumed to be prioritized over the green obstacle. The second diagram in Fig 11 corresponds to the case where no relaxation variable is used, and the end-effector avoids both obstacles by following a path equidistant from them. This was also the case when $\beta$ value was kept very large. In contrast, the third diagram in Fig. 11 incorporates a relaxation variable for the CBF constraint of the green obstacle. In this case, the path followed by the end-effector is closer to the green obstacle and farther from the red one. Additionally, by adjusting the upper bound of the relaxation variable,

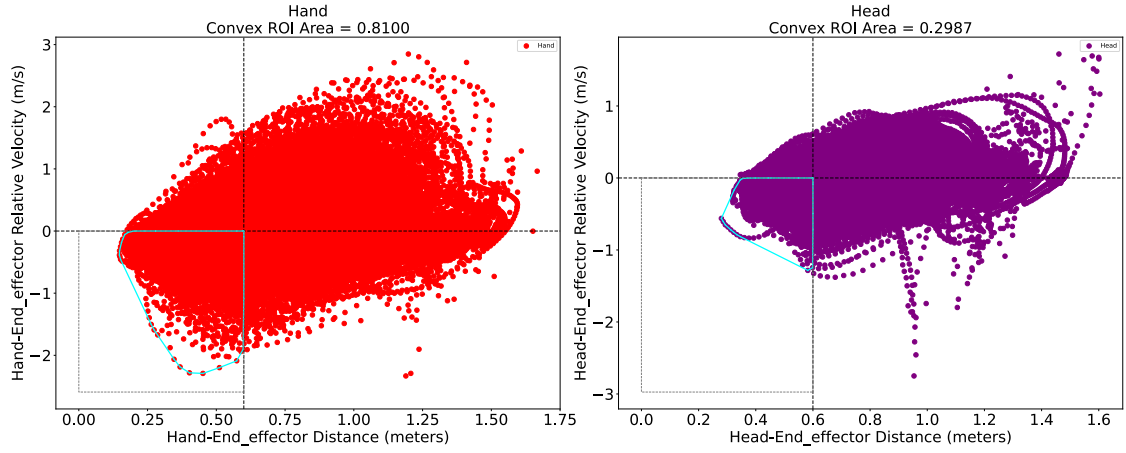

**Figure 10:** Relative velocity vs. distance from the end-effector for head and hand (40 runs). ROI ($x < 0.6$, $y < 0$) highlights close, negative-velocity interactions. The hand spans a larger convex hull (0.8100 vs. 0.2987), indicating more aggressive motion and supporting head prioritization for safety.

**Table 1**
The results show the average relative distances (meters) and velocities(meters/seconds) of the end-effector with respect to the hand and head, along with the maximum relaxation variable values reached during the experiment for four different trajectories.

| S.No | Pick & Place | | | | | Circular X-Y | | | | | Circular Y-Z | | | | | Square | | | | |
|---|---|---|---|---|---|---|---|---|---|---|---|---|---|---|---|---|---|---|---|---|
| | Head | | Hand | | relax | Head | | Hand | | relax | Head | | Hand | | relax | Head | | Hand | | relax |
| | $d_{rel}$ | $v_{rel}$ | $d_{rel}$ | $v_{rel}$ | $\delta_{max}$ | $d_{rel}$ | $v_{rel}$ | $d_{rel}$ | $v_{rel}$ | $\delta_{max}$ | $d_{rel}$ | $v_{rel}$ | $d_{rel}$ | $v_{rel}$ | $\delta_{max}$ | $d_{rel}$ | $v_{rel}$ | $d_{rel}$ | $v_{rel}$ | $\delta_{max}$ |
| 1 | 0.759 | -0.113 | 0.655 | -0.136 | 0.253 | 0.711 | -0.170 | 0.579 | -0.136 | 0.337 | 0.659 | -0.121 | 0.605 | -0.149 | 0.398 | 0.719 | -0.143 | 0.570 | -0.149 | 0.720 |
| 2 | 0.715 | -0.160 | 0.594 | -0.213 | 0.368 | 0.711 | -0.154 | 0.617 | -0.177 | 0.309 | 0.654 | -0.130 | 0.634 | -0.157 | 0.358 | 0.642 | -0.128 | 0.553 | -0.189 | 0.722 |
| 3 | 0.732 | -0.141 | 0.633 | -0.174 | 0.479 | 0.776 | -0.227 | 0.623 | -0.242 | 0.750 | 0.717 | -0.126 | 0.622 | -0.151 | 0.556 | 0.643 | -0.146 | 0.605 | -0.196 | 0.819 |
| 4 | 0.690 | -0.164 | 0.654 | -0.209 | 0.585 | 0.723 | -0.233 | 0.589 | -0.206 | 0.588 | 0.710 | -0.143 | 0.570 | -0.096 | 0.223 | 0.679 | -0.110 | 0.515 | -0.111 | 0.346 |
| 5 | 0.798 | -0.151 | 0.804 | -0.236 | 1.011 | 0.678 | -0.179 | 0.569 | -0.137 | 0.413 | 0.692 | -0.129 | 0.576 | -0.137 | 0.457 | 0.670 | -0.117 | 0.575 | -0.154 | 0.539 |
| 6 | 0.757 | -0.146 | 0.731 | -0.159 | 0.462 | 0.733 | -0.164 | 0.601 | -0.250 | 0.447 | 0.688 | -0.083 | 0.607 | -0.132 | 0.557 | 0.634 | -0.117 | 0.529 | -0.120 | 0.467 |
| 7 | 0.758 | -0.139 | 0.622 | -0.260 | 0.877 | 0.762 | -0.161 | 0.592 | -0.148 | 0.355 | 0.679 | -0.097 | 0.635 | -0.170 | 0.527 | 0.670 | -0.111 | 0.543 | -0.114 | 0.376 |
| 8 | 0.734 | -0.095 | 0.726 | -0.127 | 0.443 | 0.747 | -0.158 | 0.555 | -0.136 | 0.593 | 0.687 | -0.109 | 0.640 | -0.163 | 0.490 | 0.591 | -0.082 | 0.628 | -0.112 | 0.196 |
| 9 | 0.806 | -0.082 | 0.707 | -0.131 | 0.329 | 0.747 | -0.133 | 0.520 | -0.074 | 0.232 | 0.762 | -0.138 | 0.699 | -0.208 | 0.654 | 0.614 | -0.116 | 0.528 | -0.127 | 0.671 |
| 10 | 0.754 | -0.089 | 0.629 | -0.139 | 0.704 | 0.689 | -0.115 | 0.589 | -0.187 | 0.372 | 0.774 | -0.108 | 0.687 | -0.173 | 0.541 | 0.656 | -0.098 | 0.601 | -0.089 | 0.430 |

the allowed minimum distance between the lower-priority obstacle (green) and the end-effector can be controlled. If the upper bound is set too high, the end-effector may collide with the lower-risk priority obstacle (green), as shown in the fifth diagram in Fig. 11. Thus, the relaxation variable provides a tunable mechanism to enforce safety prioritization among competing constraints in close-proximity scenarios. The velocity of the obstacle is calculated using a spring-damper system:

$$\mathbf{v}_{\text{desired}} = k(\mathbf{p}_{\text{EE}} - \mathbf{p}_{\text{obs}}) - b\,\mathbf{v}_{\text{obs}} \quad (17)$$

where $k$ is the spring constant, $b$ is the damping coefficient and $\|\mathbf{v}_{\text{desired}}\|$ is shortened to a maximum speed specific to the scenario. To evaluate the effectiveness of the proposed Hierarchical-CBF control strategy, we conducted comparative experiments in three scenarios of increasing difficulty, Easy, Medium and Hard, characterized by varying velocities of the obstacle approach and initial clearances shown in Table 3. The exact metrics used to define the conditions for the easy, medium, and hard scenarios are detailed in Table 2. Three spherical obstacles were randomly introduced in each scenario, each assigned a distinct priority:

**Table 2**
Scenario parameters for obstacle motion dynamics.

| Parameter | Easy | Medium | Hard |
|---|---|---|---|
| Initial distance (m) | [0.75, 1.2] | [0.6, 0.9] | [0.3, 0.6] |
| Spring constant $k$ | 0.5 | 2.0 | 4.0 |
| Damping coefficient $b$ | 1.0 | 0.2 | 0.1 |
| Max speed (m/s) | 0.05 | 0.10 | 0.15 |

red (highest), blue (medium), and green (lowest). For all relaxation-enabled scenarios, the CBF gains were fixed at $\gamma_{\text{red}} = \gamma_{\text{green}} = \gamma_{\text{blue}} = 1.0$. Relaxation weights were set at $\beta_{\text{green}} = 250$, $\beta_{\text{blue}} = 500$, and performance weight (penalizing deviation from the desired velocity) was set to 200. The number of constraints involved was 25μ7 for limiting the lower and upper bounds of each joint velocity, and 18 safety CBF certificates for the last 6 links and 3 obstacles. The Root Mean Square Error (RMSE) of trajectory tracking and the minimum distance ($d_{\min}$) between the robot and each obstacle were recorded for both the baseline CBF and the Hierarchical-CBF approaches. In the Hierarchical-CBF formulation, we report the maximum values of the relaxation variable ($\delta_{max}$) for the blue and green constraints, indicating how the system adapted to avoid infeasibility. The cells highlighted in red correspond to instances where the constraints could not be met. Notably, while the baseline CBF approach encountered constraint violations, including contacts with higher-priority obstacles, the Hierarchical-CBF controller consistently avoided collisions with the red (highest-priority) obstacle. When constraint violations occurred, they were limited to the lower priority blue or green obstacles, demonstrating the effectiveness of the prioritization scheme. These results confirm that the introduction of hierarchical relaxation enables safe and efficient task execution in environments where strict constraint satisfaction may be infeasible, particularly under dynamic and high-risk conditions.

**Table 3**
Comparison of CBF and Hierarchical-CBF in Easy, Medium, and Hard scenarios with varying obstacle clearance and speed. Obstacles (red, blue, green) denote decreasing priority. RMSE, $d_{min}$, and $\delta_{max}$ are reported; red cells indicate constraint violations. Hierarchical-CBF maintained safety for high-priority obstacles.

| S.No | Without Relaxation Variable | | | | | | With Relaxation Variable | | | | | | | | | | | |
|---|---|---|---|---|---|---|---|---|---|---|---|---|---|---|---|---|---|---|
| | Easy | | Medium | | Hard | | Easy | | | | Medium | | | | Hard | | | |
| | RMSE | $d_{min}$ | RMSE | $d_{min}$ | RMSE | $d_{min}$ | RMSE | $d_{min}$ | $\delta_{max}$ | $\delta_{max}$ | RMSE | $d_{min}$ | $\delta_{max}$ | $\delta_{max}$ | RMSE | $d_{min}$ | $\delta_{max}$ | $\delta_{max}$ |
| 1 | 0.2452 | 0.2491 | 0.3763 | 0.1671 | 0.5456 | 0.0838 | 0.46 | 0.241 | 0.0014 | 0.0212 | 0.559 | 0.123 | 0.345 | 0.271 | 0.6659 | 0.1046 | 0.091 | 0.169 |
| 2 | 0.2654 | 0.2537 | 0.35 | 0.20 | 0.3398 | 0.1047 | 0.64 | 0.156 | 0.0 | 0.030 | 0.409 | 0.1137 | 0.268 | 0.277 | 0.769 | 0.105 | 0.08 | 0.228 |
| 3 | 0.245 | 0.247 | 0.3210 | 0.1631 | 0.0464 | 0.1905 | 0.275 | 0.243 | 0.0033 | 0.0089 | 0.714 | 0.129 | 0.0276 | 0.2432 | 0.679 | 0.0921 | 0.0571 | 0.564 |
| 4 | 0.4067 | 0.2572 | 0.1524 | 0.1933 | 0.5547 | 0.0812 | 0.406 | 0.203 | 0.019 | 0.074 | 0.697 | 0.119 | 0.038 | 0.2288 | 0.7398 | 0.0763 | 0.29 | 0.30 |
| 5 | 0.2575 | 0.2246 | 0.3377 | 0.1920 | 0.6028 | 0.1209 | 0.56 | 0.169 | 0.120 | 0.0671 | 0.52 | 0.103 | 0.126 | 0.252 | 0.720 | 0.105 | 0.1056 | 0.1920 |
| 6 | 0.3259 | 0.2477 | 0.5866 | 0.20 | 0.5726 | 0.1741 | 0.35 | 0.14 | 0.0172 | 0.1157 | 0.281 | 0.101 | 0.077 | 0.179 | 0.63 | 0.106 | 0.197 | 0.4683 |
| 7 | 0.3448 | 0.2553 | 0.5352 | 0.2082 | 0.6664 | 0.1401 | 0.38 | 0.192 | 0.0131 | 0.128 | 0.447 | 0.1707 | 0.0 | 0.025 | 0.739 | 0.1159 | 0.0850 | 0.2057 |
| 8 | 0.5553 | 0.2181 | 0.3328 | 0.2239 | 0.5074 | 0.1245 | 0.549 | 0.202 | 0.0083 | 0.037 | 0.728 | 0.146 | 0.184 | 0.071 | 0.751 | 0.102 | 0.173 | 0.208 |
| 9 | 0.2627 | 0.2575 | 0.3542 | 0.1768 | 0.0786 | 0.1791 | 0.54 | 0.1483 | 0.051 | 0.108 | 0.710 | 0.166 | 0.0468 | 0.0244 | 0.627 | 0.037 | 0.125 | 0.0269 |
| 10 | 0.375 | 0.2464 | 0.1879 | 0.2073 | 0.0865 | 0.2054 | 0.421 | 0.237 | 0.0115 | 0.0118 | 0.659 | 0.136 | 0.1030 | 0.1760 | 0.726 | 0.108 | 0.189 | 0.366 |

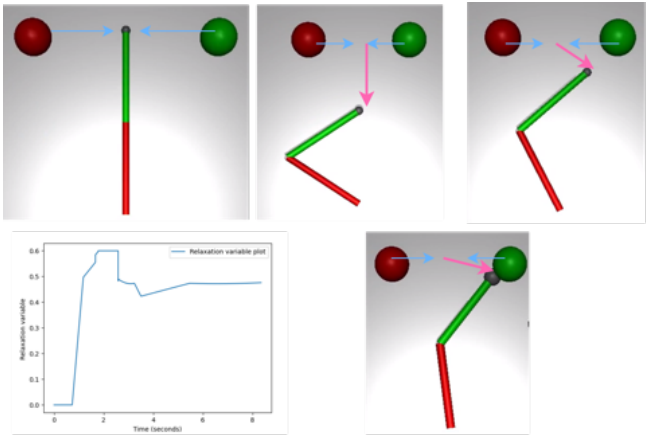

**Figure 11:** Simulation of a 2D manipulator illustrating the effect of the relaxation variable. Diagrams show: (1) initial setup, (2) no relaxation, (3) red obstacle prioritized over green, (4) relaxation variable capped at 0.6, and (5) high upper bound on relaxation.

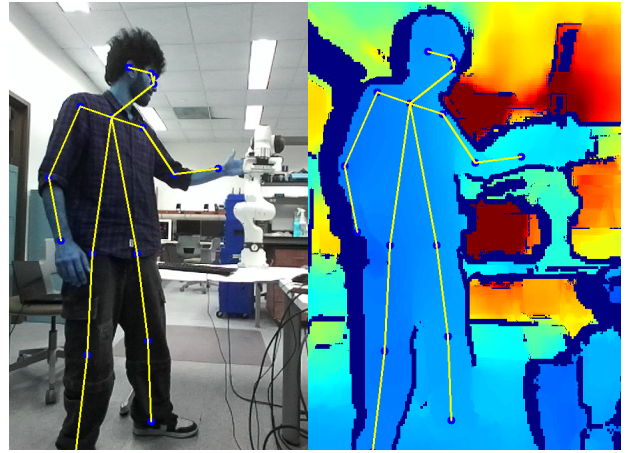

**Figure 12:** Human keypoint detection with ZED 2i. *Left*: Skeleton on RGB stream. *Right*: 3D keypoints on depth map showing spatial localization.

### 4.2. Human Body Detection and Tracking with the ZED 2i Camera

To attain accurate spatial awareness of the human co-worker in shared workspaces, we use the ZED 2i stereo camera (Stereolabs) with their official body tracking SDK Stereolabs (2025). The camera is initialized at HD720 resolution with 60 FPS frame rate, using performance depth mode. The HUMAN_BODY_ACCURATE model is used for keypoint detection, providing high-precision estimates of 34 skeletal joints, including head, torso, arms, hands, and legs. The body tracking setting includes frame-to-frame tracking and body fitting to improve joint localization. Segmentation is disabled to reduce computation. The detection confidence threshold is set to 52, suitable for indoor environments with a balance of reliability and robustness. The ZED 2i offers a depth error of 1% up to 3 meters and a depth error of less than 5% up to 15 meters, with a wide horizontal field of view of 120- Abdelsalam, Mansour, Porras, and Happonen (2024). Fig. 12 illustrates an example of the RGB overlay with tracked keypoints. The 3D keypoints that were extracted, namely head and hands, are used to dynamically construct CBF constraints in real-time for proactive safety enforcement in the human-robot interaction close to proximity.

## 5. Additional Analysis

Using a hierarchical stability analysis, considering Lyapunov convergence and the boundedness of the signals, we can prove that the controller proposed in Fig. 5 is stable. Initially, we prove that the controller is stable and independent of the safety layer. Then, we prove that the safety layer only ensures that the robot is operated safely in a work space when a human is in close proximity to the robot and will not affect the stability of the controller. Afterward, we provide a complexity analysis of the proposed safe controller approach followed by risk assessment.

### 5.1. Stability Analysis

The closed-loop error dynamics in Eq. (1) before the hierarchical CBF is applied can be rewritten as

$$e = p_d - p_e \qquad , \dot{e} = \dot{p}_d - \dot{p}_e. \qquad (18)$$

Using $\dot{q}_{perf}$ defined in Eq. (5), $\dot{e} = \dot{p}_d - J_{ee}(q)\dot{q}_{perf}$ and putting them in Eq. (18), a closed loop form of the control system for a single integrator layer can be written as

$$\dot{e} + \lambda e = 0 \qquad (19)$$

where $\lambda > 0$ means that the error $e$ will converge to zero. Quantitatively, the robot tracking error for any given time can be treated within the linearized approximation at a rate given by $|e(t)| \simeq e^{-\lambda t}|e(0)|$. The performance controller ensures that the velocity commands are followed with a negligible tracking error by choosing a sufficiently large $\lambda$ value in Eq. (5). The robot is equipped with the Franka Control Interface (FCI), a low-level torque and speed control interface for joints that leverages the available Lagrangian Dynamic Robot Model Haddadin, Parusel, Johannsmeier, Golz, Gabl, Walch, Sabaghian, Jähne, Hausperger, and Haddadin (2022). The higher level of the proposed hierarchical safe controller replaces $\dot{q}_{perf}$ with $\dot{q}_{safe}$ by solving the nonlinear constrained optimization problem in (15) and the final closed loop form can be written as

$$\dot{e} + \lambda e = \zeta \qquad (20)$$

where $\zeta$ is caused by $\dot{q}_{safe} - \dot{q}_{perf}$ and is bounded as $\dot{q}_{min} < \dot{q}_{safe}, \dot{q}_{perf} < \dot{q}_{max}$. Hence, based on the Bounded-Input, Bounded-Output (BIBO) stability theory, this ensures that if an input is bounded, the output remains bounded. Therefore, the tracking errors $e$ asymptotically converge to a compact set $S = \{e(t) : |e(t)| \leq \frac{\sqrt{2}\lambda_g}{\sqrt{\lambda_a \lambda_{\min}(P)}}\}$ and the control system is stable, and the CBF solver ensures that the state of the system remains in safe subsets within $\mathcal{X}_s$.

### 5.2. Complexity Analysis of Solver

The performance controller has a complexity of $O(n^3)$ (mainly due to the Jacobian pseudo-inverse), while CBFs exhibit $O(n^{3.5})$ (driven by LDLT Cholesky decomposition inside the ProxSuite solver Bambade et al. (2023)). Since these components are used sequentially in the proposed control approach, the net complexity will be the sum of the individual complexities. However, since $O(n^{3.5})$ dominates $O(n^3)$, the overall complexity will be governed by the term with the highest exponent: $O(n^{3.5})$. The proposed control approach is based on the detected locations of human keypoints, which are updated at a frequency of $60Hz$. As a result, the computation time of the solver does not significantly impact the overall performance. During the experiments, it was observed that each control cycle took an average of $16.6ms$, and the solver itself only required $0.16ms$ milliseconds to generate the optimal solution. For the operation of ZED2i Camera, Nvidia GeForce RTX 4060 Ti graphics card was used, while the solver computation was performed on 13th Gen Intel Core i7-13700kfx24 computer system.

## 6. Conclusion

In this study, we developed and implemented a CBF-based proactive hierarchical control strategy for an autonomous industrial robot manipulator operating in close proximity to human workers. By integrating real-time human pose detection using a depth camera, the system was able to proactively prioritize safety constraints - specifically assigning higher importance to more sensitive body parts, such as the head over the hand. The experimental results showed that the robot could rapidly adapt its trajectory in response to the presence of humans, effectively balancing safety and efficiency while reducing the risk of injury. These findings demonstrate the feasibility of using CBFs to dynamically adjust control policies in real time to mitigate the risk of collision and serious injury. In general, this work advances safety protocols for collaborative robotic systems and can be extended to a wide range of autonomous and mobile robots, particularly in scenarios where human proximity and prioritized risk assessment are critical to minimize harm. In future work, we aim to explore a learning-based strategy for adaptively tuning the $\beta$ values in real time based on obstacle semantics and motion characteristics. This would enhance the robot's ability to balance safety and collaboration in complex, dynamic environments.

## Declaration of Generative AI and AI-assisted technologies in the writing process

During the preparation of this work, the authors used OpenAI ChatGPT to improve the clarity and grammar of the text. After using this tool, the authors reviewed and edited the content as needed and assume full responsibility for the content of the publication.

## Acknowledgment

The authors express their sincere gratitude to Professor Katsuo Kurabayashi and Dr. Rui Li of New York University

for their invaluable guidance and support throughout this project. Their expertise and encouragement were instrumental to the success of this work. We also extend our heartfelt thanks to Shaman Mananbhai Panchal and Gopikrishnan Nair for their assistance and collaboration during the Master's project course at NYU, which greatly contributed to the development of this research.

# CRediT authorship contribution statement

**Patanjali Maithani:** Writing – review & editing, Writing – original draft, Visualization, Validation, Methodology, Investigation, Formal analysis, Data curation, Conceptualization. **Aliasghar Arab:** Writing – review & editing, Validation, Methodology, Supervision, Investigation, Conceptualization. **Farshad Khorrami:** Writing – review & editing, Supervision, Investigation, Conceptualization. **Prashanth Krishnamurthy:** Writing – review & editing, Supervision, Investigation, Conceptualization.